\documentclass[letter,oneside]{article}
\usepackage[margin=1.0in]{geometry}

\usepackage{xcolor}

\usepackage[utf8]{inputenc} 
\usepackage[T1]{fontenc}    
\usepackage{hyperref}       
\usepackage{url}            
\usepackage{booktabs}       
\usepackage{amsfonts}       
\usepackage{nicefrac}       
\usepackage{microtype}      
\usepackage{graphicx}
\usepackage{algorithmic,amsmath,amssymb,amsthm}
\usepackage[ruled, vlined]{algorithm2e}
\usepackage{multicol}
\usepackage{multirow}
\usepackage{listings}
\usepackage{xcolor}

\definecolor{codegreen}{rgb}{0,0.6,0}
\definecolor{codegray}{rgb}{0.5,0.5,0.5}
\definecolor{codepurple}{rgb}{0.58,0,0.82}
\definecolor{backcolour}{rgb}{0.95,0.95,0.92}

\lstdefinestyle{mystyle}{
    backgroundcolor=\color{backcolour},   
    commentstyle=\color{codegreen},
    keywordstyle=\color{magenta},
    numberstyle=\tiny\color{codegray},
    stringstyle=\color{codepurple},
    basicstyle=\ttfamily\footnotesize,
    breakatwhitespace=false,         
    breaklines=true,                 
    captionpos=b,                    
    keepspaces=true,                 
    numbers=left,                    
    numbersep=5pt,                  
    showspaces=false,                
    showstringspaces=false,
    showtabs=false,                  
    tabsize=2
}

\newenvironment{definition}[1][Definition]{\begin{trivlist}
\item[\hskip \labelsep {\bfseries #1}]}{\end{trivlist}}

\allowdisplaybreaks

\DeclareMathOperator*{\argmax}{argmax}

\newtheorem{lemma}{Lemma}

\title{Multihop: Leveraging Complex Models to Learn Accurate Simple Models}

 \author{
 Amit Dhurandhar\thanks{Equal contribution.}\\
   IBM Research, NY\\
      \texttt{adhuran@us.ibm.com}
      \and
   Tejaswini Pedapati$^*$\\
   IBM Research, NY\\
   \texttt{tejaswinip@us.ibm.com}
}

\date{}
\begin{document}

\maketitle

\begin{abstract}
  Improving the performance of a low performing simple model with the assistance of a high performing complex model has been of significant interest recently as it finds applications in important problems such as explainable artificial intelligence, model compression, robust model building and learning from small data. Known approaches to this problem (viz. Knowledge Distillation, Model compression, ProfWeight, etc.) typically transfer information directly (i.e. in a single/one hop) from the complex model to the chosen simple model through schemes that modify the target or reweight training examples on which the simple model is trained. In this paper, we propose a meta-approach where we transfer information from the complex model to the simple model by dynamically selecting and/or constructing a sequence of intermediate models of decreasing complexity that are less intricate than the original complex model. Our approach can transfer information between consecutive models in the sequence using any of the previously mentioned approaches as well as work in 1-hop fashion, thus generalizing these approaches. In the experiments on real data, we observe that we get consistent gains for different choices of models over 1-hop, which on average is more than 2\% and reaches up to 8\% in a particular case. We also empirically analyze conditions under which the multi-hop approach is likely to be beneficial over the traditional 1-hop approach, and report other interesting insights. To the best of our knowledge, this is the first work that proposes such a multi-hop approach to perform knowledge transfer given a single high performing complex model, making it in our opinion, an important methodological contribution.
\end{abstract}

\section{Introduction}
In many real scenarios where interpretability is important \cite{profweight} or robustness is critical \cite{distill} we may have strict requirements on the type of model that is desired. For example, domain experts in multiple industries such as chip manufacturing or oil production might have a strong preference for decision trees \cite{profweight}, since they understand the model well and can reason its decisions especially when things go wrong. In such a scenario the ask is typically to build the best possible decision tree, rather than some other model even though it may be more accurate. Hence, having a high performing neural network by itself is of less value, unless one can leverage it to build the best possible decision tree. It is also important in such settings to build the simple model on a fixed representation such as the original input data which presumably has features that the expert can interpret. Thus, learning an intermediate representation \cite{crd} on which to train a simple model may not be desirable or even allowable. There are also other settings where transferring to a fixed (set of) model might be important such as when we have strict memory/power/hardware constraints as in the case of Unmanned Aerial Vehicles (UAVs) or edge devices in Internet of Things (IoT) kind of systems. Moreover, having an understandable model can lead many times to better real world performance with a human in the loop, rather than blindly basing ones decisions on a black-box model \cite{kushwhyInt}.

\begin{figure*}[t]
  \centering  
      \includegraphics[width=0.45\textwidth]{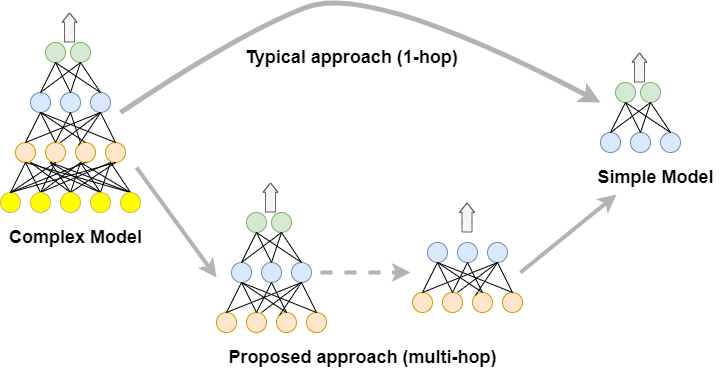}
      \includegraphics[width=0.54\textwidth]{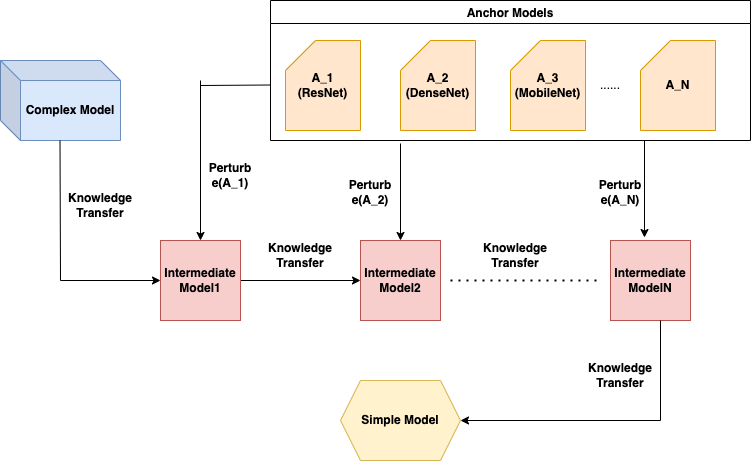}
      
  \caption{Above in the left figure we see a contrast between the typical approaches that perform knowledge transfer to simple models given a single complex model which are inexorably 1-hop, while our proposal is to generalize this to multihop with different intermediate models being created and selected that lead to superior simple model performance. In the right figure we see more details about our approach where we see a 3-hop transfer with intermediate models that formed by perturbing (possible) anchor models.}
  \label{fig:multihop}
\end{figure*}

\begin{figure}
\centering
\includegraphics[width=0.4\textwidth]{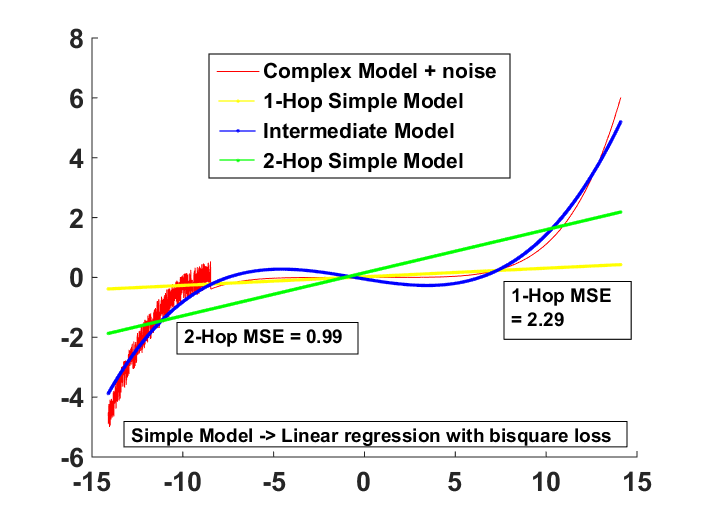}
\caption{Above we see a synthetic data example which showcases the potential of our approach. The complex model is a fifth degree polynomial and the simple model is a (robust) linear model with bisquare loss. We observe that the simple model's mean squared error (MSE) from fitting to the complex models noisy predictions is much higher than fitting the simple model to a third degree polynomial (intermediate model) which in turn was fit to the noisy complex models predictions.}
\label{fig:synegintro}
\end{figure}

In our current work, we thus provide solutions for the setting where we are given a fixed simple model and want to improve its performance based on a given complex model. Our main idea, as we will see later, is to create \emph{anchor models} from a given complex model and leverage some of these to transfer information from the complex model. Moreover, our approach also explores variations/perturbations of these anchor models where we create what we aver to as \emph{intermediate models} thus leading to an even richer model class than just the anchors themselves for more effective knowledge transfer. A key point to notice here is that the intermediate models \emph{do not} have to respect the constraints that led to the choice of the simple model (viz. interpretability or size etc.) as only the simple model will be deployed which gives us a lot of flexibility in not only designing the intermediate models, but also the number of them we use in the sequence (i.e. number of hops) to transfer information from the complex model to the simple model. An illustration of this is seen in Figure \ref{fig:multihop}, where we see the typical or traditional approach which transfers information directly, while ours can use multiple hops based on intermediate models of decreasing complexity, but which are more complex than the simple model. It is important to note that direct or 1-hop transfer is just a special case of our approach.

With the intention of motivating the idea, in Figure \ref{fig:synegintro}, we observe a synthetic data example. Here a fifth degree polynomial is the complex model and a linear model with bisquare loss is the simple model. Uniform random noise is added to the first 20\% of the complex models predictions. We see in the figure that when the simple model is directly fit to the noisy predictions of the complex model the mean squared error of the simple model is 2.29. This error is reduced by more than half when an intermediate model, which is a third degree polynomial in this case, is used. A detailed discussion of this setup is provided in the experimental section where qualitatively similar results are seen in the noiseless setting and more insight is provided about its behavior.

The main motivation for our proposed approach is that smoothing the prediction surface at the expense of some degree of fidelity seems to be useful for knowledge transfer to simple models. Knowledge distillation does this to some degree where a simple model is fit to the (soft) predictions of a complex model, which although accurate, has of course lower fidelity than the original (ground truth) training data. We generalize this idea where we choose to not limit ourselves to such a 1-hop transfer, but rather choose the optimal number of hops (and intermediate models) for further improving performance. A better tradeoff between fidelity and smoothness seems to be achieved by our methodology for knowledge transfer as can be witnessed in the experiments. Another more subtle motivation for our approach is that there is typically a loss mismatch between how models are trained and how they are evaluated. For instance, neural networks typically optimize cross-entropy loss, however, the test performance is evaluated with 0-1 loss. In such cases of loss mismatch, which is also seen in our synthetic experiment in Figure \ref{fig:synegintro}, the multihop approach seems to have even more promise, as smoother but less faithful models w.r.t. one loss can transfer better relative to a different loss. This subtle, but interesting aspect may inspire more methods and deeper analysis in the future.

\section{Related Work}

As described before there are multiple approaches to transfer information from a complex to a simpler model.
Knowledge Distillation (KD) \cite{distill,distillnew,priv16} is arguably the most popular approach for doing this where the simple model is typically a neural network. It entails minimizing the cross-entropy loss of a simpler network based on calibrated confidences \cite{calib} of a more complex network accomplished by temperature scaling or by minimizing Kullback-Liebler (KL) divergence. The simple networks are typically of the same (or similar) depth but thinned down \cite{fitnet}. Other strategies such as model compression \cite{modelcompr,modelcompr2,bastani2017interpreting} are also conceptually similar to KD, where the actual outputs are replaced by predictions from the complex model. In \cite{dehghani2017fidelity}, authors use soft targets and their uncertainty estimates to inform a student model on a larger dataset with more noisy labels. Uncertainty estimates are obtained from Gaussian Process Regression done on a dataset that has less noisy labels. In \cite{furlanello2018born}, authors train a student neural network that is identically parameterized to the original one by fitting to soft scores rescaled by temperature.  \cite{frosst2017distilling} define a new class of decision trees called soft decision trees to enable it to fit soft targets (classic KD) of a neural network. 
The weighting based approaches \cite{profweight,sratio} on the other hand do not alter the target to which the simple model is fit, rather learn an importance weighting of the training inputs on which the simple can be trained. These procedures can be easily applied to models optimizing losses other than cross-entropy such as hinge loss or squared loss also providing interpretation of which inputs are more (or less) important. \cite{ren2018learning} propose to weight samples to make deep learning robust by tuning the weights on a validation set through gradient descent. This is different from our problem where we have a pre-trained complex model and wish to improve a given simple model. As such, all these approaches transfer information directly (i.e. in 1-hop) from a complex to a simple model. Even the recent approach that learns new representations \cite{crd}, which may not be suitable for our problem setting as mentioned earlier, does so directly from the complex model and hence belongs to the 1-hop category.

There are also approaches which transfer information from multiple complex models to one or more simple models. In \cite{twostageMC}, multiple complex models are trained and multiple simple models are distilled. These simple models are then ensembled to form the final model. The ensembling can of course break the requirements for applications motivated in this paper such as interpretability or deployment in resource constrained settings. Another approach \cite{takd} transfers information in multihop fashion but assumes access to multiple complex models of decreasing complexity termed as teaching assistants, unlike ours where we just have access to a single complex model and show how to create anchor/intermediate models. The method also is limited to transferring information between neural networks with the same architecture, but differing depths and using KD. Other transfer strategies such as weighting based are not suited for it. Ours can work for arbitrary complex-simple model combinations and also with transfer methods other than KD. Moreover, given $k$ anchor models and a maximum of $m$ hops the method has time complexity of $O(mk^2)$ as opposed to $O(mn)$ where $n\le k$ which is ours. Nonetheless, we compare with an enhanced version of this method in the experiments and show consistently better results.



\section{Methodology}
\label{sec:meth}

\subsection{Problem Formulation}

Let $X\times Y$ denote the input-output space. For a positive integer $d$ let $C:\mathbb{R}^d\rightarrow \mathbb{R}$ denote the complex model and $\mathcal{S}:\mathbb{R}^d\rightarrow \mathbb{R}$ the class of simple models. Let $t(.,.,.)$ be a function that takes the complex model, simple model class and intermediate models as input and returns a particular simple model as output. Procedures such as KD or ProfWeight (PW) could be used to transfer information in each hop. Assume $k$ anchor models $\mathcal{A}_1,...,\mathcal{A}_k$ all mapping $\mathbb{R}^d\rightarrow \mathbb{R}$ of complexity lower than $C$. Examples of how to create these from different commonly used complex models $C$ are described in section \textit{Anchor Models}. Further assume a randomized perturbation function $e(.)$ which takes as input an anchor model and outputs another model with same or "slightly" different architecture -- i.e. a potential intermediate model is available. We also show example instantiations of this for popular models in section \textit{Intermediate Models} . With this and given a likelihood or reward function $\rho(.,.)$, we want to optimize the following objective over a dataset $D$ where knowledge transfer up to $m$ hops is allowed:
\begin{align}
    S^*= \argmax\limits_{S\in \{t(C,\mathcal{S},\mathcal{I})| \mathcal{I}\in  \{e(\mathcal{A}_i)|i=0,...,k)\}^m\}} \frac{1}{|D|}\sum_{(x,y)\in D}\rho(S(x),y)
\end{align}
where we define $e(\mathcal{A}_0)=\phi$ that is absence of a (intermediate) model or no model, 
\begin{align*}
&\{e(\mathcal{A}_i)|i=0,...,k\}^m\\&= \underbrace{\{e(\mathcal{A}_i)|i=0,...,k\}\otimes ...\otimes\{e(\mathcal{A}_i)|i=0,...,k\}}_m
\end{align*}
that is a m-ary cartesian product of the set of intermediate models which are created from the anchor models and $|.|$ denotes cardinality.

Loosely speaking, we want to find the combination of intermediate models derived from the anchor models that transfers information from the complex models the best so as to obtain a high performing simple model. There are a few things to note about this formulation. First, direct or 1-hop transfer is just a special case of this formulation where $e(\mathcal{A}_0)$ or $\phi$ is selected at each hop. Second, the $e(.)$ functions are randomized and so can lead to different intermediate models at different hops even for the same anchor model making the search space more complicated but also richer. One could of course set $e(.)$ to an identity function in which case intermediate models would be specific anchor models and the problem becomes solely a selection problem. Third, the search space for the best combination is exponential in $m$ and hence methods that intelligently explore promising parts of the space have to be developed. Even for small $m$ and $k$ brute-force search can still be expensive as each operation requires training models that are produced by perturbing the anchor models.

\begin{algorithm}[t]
    \caption{Multihop Stochastic Transfer Method (MSTM). 
    }
    \label{alg:multihopalgo}
\begin{algorithmic}
\STATE \textbf{Input:} Complex model $C$, dataset $D$, maximum hops $m$, transfer function $t(.,.,.)$ and simple model class $\mathcal{S}$.
\STATE
\STATE Create $k$ anchor models $\mathcal{A}_1,...,\mathcal{A}_k$ with decreasing complexity from $C$ \COMMENT{See section \textit{Anchor Models}}
\STATE Use appropriate perturbation function $e(.)$ based on Algorithm \ref{alg:intermalgo} \COMMENT{See section \textit{Intermediate Models}}
\STATE Initialize $I \leftarrow \{1,...,k\}$, $i\leftarrow 1$ and $\mathcal{I}\leftarrow \phi$

\WHILE{$i\le m$}
\STATE Randomly choose $n\le |I|$ elements from $I$ along with 0 and assign to set $I_{[n]}$
\STATE $\mathcal{I}_i\leftarrow\argmax\limits_{e(\mathcal{A}_j)|j\in I_{[n]}} \sum_{(x,y)\in D}\rho\left(t(C,\mathcal{S},\mathcal{I}\cup e(\mathcal{A}_j))(x),y\right)$
\STATE $I_i\leftarrow j$ \COMMENT{index of the anchor model $\mathcal{A}_j$ chosen above to maximize the reward in iteration $i$}
\STATE $\mathcal{I}\leftarrow \mathcal{I}\cup\mathcal{I}_{i}$
\STATE $I\leftarrow I-\{r|r=1,...,I_i\}$\COMMENT{Consider only lower complexity anchor models for next hop.} 
\STATE \textbf{if}{ $I==\phi$} \textbf{then} break \textbf{end if}
\STATE $i\leftarrow i+1$
\ENDWHILE
\STATE \textbf{Return} $S\leftarrow t(C,\mathcal{S},\mathcal{I})$ 
\end{algorithmic}
\end{algorithm}

\subsection{Method Description}

Given the above formulation we now propose our approach \emph{Multihop Stochastic Transfer Method} (MSTM) which is detailed in algorithm \ref{alg:multihopalgo}. It is a type of stochastic greedy algorithm, where at each hop the intermediate model that best improves the simple model is chosen based on evolving or perturbing a given set of randomly sub-selected anchor models. The perturbations can also be stochastic and non-parametric as we describe below in subsection \textit{Intermediate Models}. Moreover, to make the search more efficient, as training models is expensive, at each hop we only consider anchor models of lower complexity than the ones already chosen in the previous hops to create intermediate models. This makes intuitive sense since adding a more complicated model will in all likelihood lead to it exactly emulating the previous intermediate model as its hypothesis class would be a superset. This would bring little additional value to transferring knowledge from the original complex model. Hence, our approach can be perceived as a refined version of the typical stochastic greedy procedures.

Another thing to note is that at each hop not adding an intermediate model is allowed in our approach. Moreover, the function $t(.,.,.)$ could be instantiated based on different transfer strategies such as KD or weighting based ones, not to mention their combinations, i.e. KD for some hops and PW for others, are also possible. This makes the traditional 1-hop approaches special cases of our method.

From a theoretical standpoint we can show that even a restricted version of our algorithm can have a constant factor approximation guarantee for weakly submodular loss functions \cite{weaksubInit} such as generalized least squares, likelihood functions of exponential families like Gaussian, Bernoulli, Dirichlet and any arbitrary M-estimator which a simple model (e.g. linear regression, logistic regression) may very well optimize, under the intuitive assumption that if needed, a more complex intermediate model can exactly emulate a simpler one. Before we formally state our result we define the concept of submodularity ratio and consequently weak submodularity.

\begin{definition}\cite{weaksubInit}
Let $L, P \subset \{1,...,q\}=[q]$ be two disjoint sets, and $f:2^{[q]}\rightarrow R$. The submodularity ratio of $L$ w.r.t. $P$ is given by: $\gamma_{L,P} = \frac{\sum_{i\in P}\left(f(L\cup i)-f(L)\right)}{f(L\cup P)-f(L)}$
\end{definition}
The function $f(.)$ is submodular if and only if $\forall L,P$, $\gamma_{L,P}\ge 1$. However, if $\gamma_{L,P}$ can be shown to be bounded away from 0 $\forall L,P$, but not necessarily $\ge 1$, then $f(.)$ is said to be weakly submodular. With this we now state our result:
\begin{lemma}
\label{lem1}
Given a transfer function $t(.,.,.)$ and if $e(.)$ is an identity function, where $\mathcal{A}_1,...,\mathcal{A}_k$ are anchor models with hypothesis classes $\mathcal{H}_1,...,\mathcal{H}_k$ respectively such that $\mathcal{H}_1\supset...\supset \mathcal{H}_k$, $n=\frac{k}{m}\log\left(\frac{1}{\delta}\right)$ the number of randomly selected candidate models at each iteration for some $\delta$ and the reward function $\rho(.,.)$ is weakly submodular with submodularity ratio $\gamma$, then MSTM has an approximation guarantee of $\left(1-e^{-\gamma_{\mathcal{I},m}}-\delta\right)$ for a $m$ hop transfer.
\end{lemma}
\begin{proof}[Proof Sketch]
The proof is based off on three key observations. First, the non-negative weighted sum of weakly submodular functions is weakly submodular \cite{weaksubprop}. Second, at every hop we are ignoring anchor models of equal or higher complexity than the intermediate models chosen at the previous hop. Note that the intermediate models are also anchor models since, $e(.)$ is set to be identity. Given that higher complexity implies the hypothesis class of the model is a superset of something with lower complexity, it should be able to exactly reproduce the optimal hypothesis of the simpler intermediate model given the same data. This implies ignoring such more complex models in our selection process at the current hop is admissible as this will not improve $\rho$. Consequently, our approach emulates stochastic greedy selection as we choose the best candidate amongst the $n$ randomly sub-selected ones at each hop. Third, given that our method emulates stochastic greedy the stated guarantee applies from \cite{weaksubstocastic}.
\end{proof}
Hence, for a rich class of reward functions our approach has strong performance guarantees and as we will see later it performs well empirically for even others.

\subsubsection{Anchor Models}
\label{AM}
As we have seen anchor models are an integral part of our approach. In this subsection, we show how one can create these models from many commonly used complex models.

1) \textbf{Deep Neural Networks (DNN):} For complex models that are DNNs we can create anchor models by replicating the architecture but to lesser depths. For example, if ResNet-50 is the complex model, then anchor models could be ResNet-40, ResNet-30 and so on. If we have a DenseNet-169, then we could create DenseNets with fewer layers. Similar strategies could be pursued for other deep architectures.
Moreover, well known simpler architectures such as ShuffleNet and MobileNet  can also serve as anchor models for more complicated architectures.
 
2) \textbf{Tree Ensembles:} For complex models such as boosted trees or random forests one can also easily create anchor models. In this case, the anchor models could be ensembles with fewer trees. For example, if 100 boosted trees is the complex model, then 50 boosted trees or a 50 tree random forest could be an anchor model.
    
3) \textbf{Other types of Models:} For non-ensemble complex models such as generalized linear models one could create anchor models by taking different order Taylor or Pade approximations of the functions. One could also create anchor models by doing function decompositions based on binary, ternary and higher order interactions \cite{molnardecomp}.

Thus, as can be seen above not having anchor models readily provided is not a bottleneck to using our approach as they can be created without too much difficulty for a wide range of complex models.

\begin{algorithm}[t]
\SetAlgoLined
    \caption{Design of perturbation function $e(.)$ to create intermediate models.}
    \label{alg:intermalgo}
\textbf{Input:} Anchor model $\mathcal{A}$

\uIf{$\mathcal{A}$ is a neural model}{

 num-sub-blocks = compute-subblocks($\mathcal{A}$) /*Returns number of sub-blocks in $\mathcal{A}$*/

 intermediate-model $\mathcal{M}$ = $\mathcal{A}$, choose number of perturbations $L$ and cur-step = 0

\While{\text{cur-step} $<$  $L$}{
   
   subblock-num = pick-subblock(1,num-sub-blocks) /*Randomly pick a sub-block*/
   
   is-delete = delete-or-modify-subblock() /*Randomly decide if to delete or modify*/
   
   \uIf{is-delete}{
   
   $\mathcal{M}$ = delete-subblock-from-model($\mathcal{M}$, subblock-num) /*Delete sub-block*/
     
      }\Else{
        
     $\mathcal{M}$ = perturb-subblock($\mathcal{M}$, subblock-num) /*Randomly perturb hyperparameters in the sub-block*/
    
    }
 cur-step += 1
}
}
\uElseIf{$\mathcal{A}$ is tree ensemble}{
$\mathcal{M}=$ perturb-tree-depth($\mathcal{A}$) /*For tree ensembles perturb tree depth*/
}\uElseIf{$\mathcal{A}$ is GLM}{
$\mathcal{M}=$ perturb-sparsity($\mathcal{A}$) /*For regularized GLMs change degree of sparsity*/
}
\textbf{end}

\textbf{Return} $M$
\end{algorithm}
\begin{figure*}[t]
  \centering  
      \includegraphics[width=0.33\textwidth]{2-stagesynthetic.png}
      \includegraphics[width=0.32\textwidth]{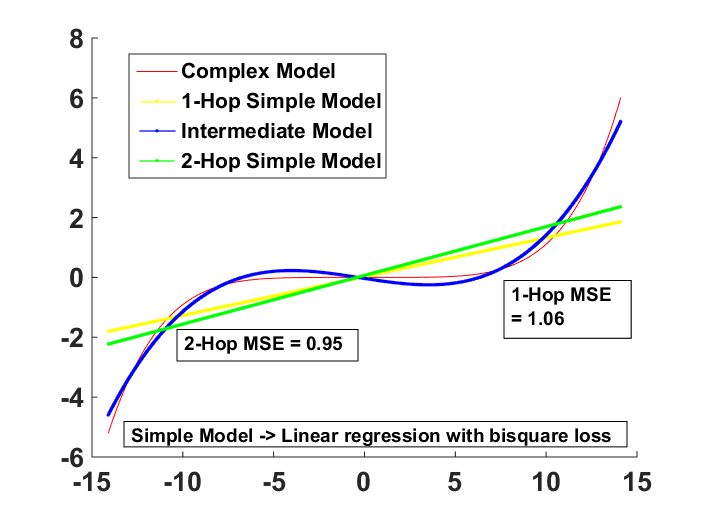}
      \includegraphics[width=0.33\textwidth]{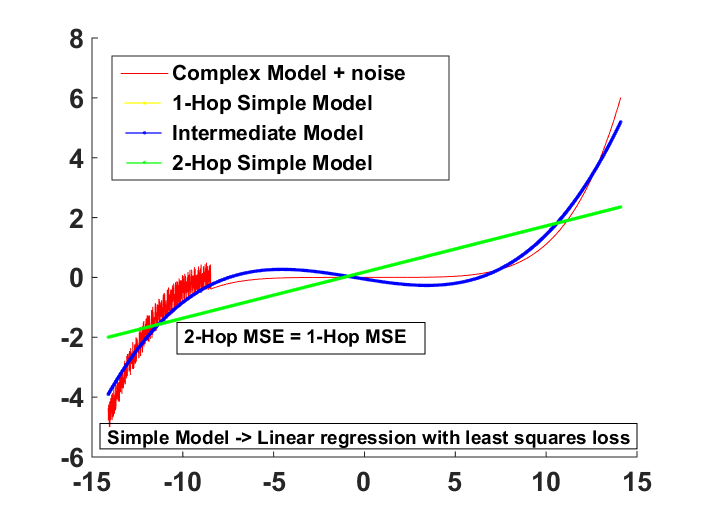}
  \caption{In the figure above we see a synthetic data example where the complex model is a fifth degree polynomial and the simple model is a (robust) linear model with bisquare loss in the left and center figures, while it is a linear model with least squares loss (i.e. polynomial of degree of 1) in the right figure. We observe that the simple models mean squared error (MSE) from fitting to the complex model with and without noise (left and center figures respectively) with bisquare loss is much higher than fitting the simple model to a third degree polynomial (intermediate model) which in turn was fit to the noisy/noiseless complex models predictions based on least squared loss. The right figure shows that when the simple model is trained using least squared loss which is the same as the evaluation loss (i.e. MSE), then both approaches perform similarly.}
  \label{multihop2}
\end{figure*}
\subsubsection{Intermediate Models} \label{IM}

We now describe how the perturbation function $e(.)$ can be designed for popular architectures to create intermediate models from anchor models. Of course, if our search space is rich enough just based on the anchor models, we could set $e(.)$ to identity in which case the (possible) intermediate models would just be some of the anchor models. Most modern neural architectures \cite{resnet,mobilenet,densenet} are formed by stacking several blocks each of which consists of repeated sub-blocks and corresponding hyperparameters that perform a certain functionality. Here are some examples of sub-blocks and hyperparameters for popular architectures:

\noindent\textbf{DenseNet \cite{densenet}:} A DenseNet has a Convolution-Relu-BatchNorm sub-block. The number of times a Convolution-Relu-BatchNorm sub-block should be repeated along with the number of channels also called growth rate are hyperparameters that determine the DenseNet architecture.


\textbf{MobileNetV2 \cite{mobilenet}:} A MobileNetV2 has residual sub-blocks which includes a depth-wise convolution layer. Each residual sub-block has stride (1 or 2), expansion factor $\alpha$, number of filters and number of times a sub-block is repeated as the hyperparameters. The depth-wise convolution expands the number of filters of the convolution by an expansion factor $\alpha$, applies the convolution function and then projects it back to the original number of filters.


\noindent\textbf{ResNet \cite{resnet}:} The sub-blocks in Residual Networks are comprised of  Convolution-BatchNorm-Relu layers. The hyperparameters are number of filters and number of times a block is repeated. 

Given instantiations of the above architectures as anchor models we can create intermediate models by (slightly) modifying their architecture as given in algorithm \ref{alg:intermalgo}.\footnote{For MLP sub-blocks would just imply layers.} The algorithm also mentions how one could create intermediate models for non-neural architectures such as tree ensembles and regularized generalized linear models (GLMs).


\section{Experiments}
\label{sec:exp}
We now perform experiments with synthetic and real data showcasing the power of our approach. Besides just reporting performance we also study where and when our approach is likely to exhibit the most benefit. We also provide additional insights to better understand our approach.

 \subsection{Synthetic Experiments}

 To understand and motivate our multihop idea we consider a simple setup where a fifth degree polynomial is the complex model and the simple model is a linear model. A trimmed down version of our method MSTM is considered, where only a single anchor model -- a third degree polynomial -- which is also the intermediate model (i.e. $e(.)$ is an identity function) is used to produce an up to 2-hop transfer.

 \noindent\textbf{Setup:} The complex model is a fifth degree polynomial with coefficients and bias $10^{-5}$, which is also the ground truth data generator. We generate input data $x$ lying between $[-14,14]$ at intervals of $0.01$ leading to roughly 2800 points. For two experiments Figure \ref{multihop2}(left) and Figure \ref{multihop2}(right) we generate uniform random noise between $[0,1]$ for the first 20\% of the inputs and add it to the predictions of the complex model. The intermediate model in all experiments is fit using least squares loss to the noisy predictions (Figure \ref{multihop2}(left) and Figure \ref{multihop2}(right)) or noiseless predictions (Figure \ref{multihop2}(center)) of the complex model (i.e. to the red curves). In the first two experiments the simple linear model is fit using bisquared loss, while in the third it is fit using least squared loss. The final evaluation of the simple model is based on mean squared error (MSE) w.r.t. the ground truth data generator i.e. the (noiseless) complex models predictions. 

 \noindent\textbf{Observations and Insights:} In Figure \ref{multihop2}(left), we see that fitting the simple model to the predictions of the intermediate model which in turn was fit to the noisy complex model produces a much better linear predictor (green line), than 1-hop fitting (yellow line). The error of the 2-hop predictor is less than half of the 1-hop predictor in this case. We see similar qualitative behavior for different degrees of noise. In fact even in the extreme case of no noise Figure \ref{multihop2}(center), we observe that the 2-hop simple model is still superior to the 1-hop fitting approach, although by a smaller margin. In Figure \ref{multihop2}(right), we see that the 1-hop approach has the same performance as 2-hop (green line overlays the yellow line) where the only difference between this and the first experiment is that the simple model is trained using least squares loss.

 The above experiments bring forth at least a couple of interesting insights. i) Smoothening of the prediction function using intermediate models seems to help knowledge transfer to simple models. ii) The evaluation loss (i.e. MSE in this case) being different from the one that the simple model optimizes can lead to different models that perform better using the multihop approach. This is a very interesting and subtle point which seems to be important to witness the power of the multihop approach over its traditional 1-hop counterpart. Note that bisquare loss has a single minima and so the improvement based on one approach but not the other cannot be attributed to serendipitously finding a better local minima. Hence, our justification based on loss mismatch has stronger grounding. 
 Moreover, such loss mismatches are prevalent in most machine learning applications as the evaluation loss is typically 0-1 loss, however, one rarely trains a model that directly minimizes it (e.g. NNs optimize cross-entropy, LR optimizes log-likelihood). Even such a simple setup 
 with limited options for our method showcases the power of our approach and helps us understand possible scenarios of when it might be useful.
\begin{table}[htbp]
\centering
\small
\caption{Public dataset characteristics, where $N$ denotes dataset size and $d$ is the dimensionality.}
\begin{tabular}{|c|c|c|c|}
  \hline
    {Dataset} & {$N$} & {$d$} & {\# of Classes}  \\
        \hline
        \hline
        {CIFAR-100} & {60K} & {$32\times32$} & {100}  \\
 \hline
     {Credit Card} & {30K} & {24} & {2}  \\
     \hline
    {Magic} & {19020} & {11} & {2}  \\
 \hline
   {Higgs} & {2K} & {28} & {2}  \\
   \hline
 {Waveform} & {5K} & {40} & {3}  \\
 \hline
{Diabetes} & {480559} & {20} & {2}  \\
 \hline
  \end{tabular}
 \label{tab:data}
\end{table}
\begin{table*}[t]
\begin{center}
\small
\caption{Below we see the \% test accuracies (Acc) of the different approaches. Best results for the simple model (based on paired t-test) are indicated in bold.
As can be seen our method MSTM is consistently better than other approaches which include direct training, 1-hop, ours with no perturbation (MSTMnp) and an enhanced version (indicated by the "+" sign) of a state-of-the-art multi-teacher approach (TAKD).}
\begin{tabular}{|c|c|c|c|c|c|c|c|c|c|}
\hline
\multirow{2}{*}{\textbf{Dataset}}& \multirow{2}{*}{\textbf{CM}} & \textbf{CM} &\multirow{2}{*}{\textbf{SM}} & \textbf{SM}& \textbf{Trans.}&\textbf{1-hop}&\textbf{MSTM}&\textbf{MSTMnp}&\textbf{TAKD+}\\
  & &\textbf{Acc} & & \textbf{Acc}&\textbf{Meth.}&\textbf{Acc}&\textbf{Acc}&\textbf{Acc}&\textbf{Acc}\\ \hline\hline
\multirow{2}{*}{CIFAR100} & \multirow{2}{*}{DNet169} & \multirow{2}{*}{79.15} & \multirow{2}{*}{LeNet} & \multirow{2}{*}{36.8} & KD & 40.64 &\textbf{42.6}& 41.05&40.49\\\cline{6-10} 
&&&&& PW & \textbf{42.49}&\textbf{42.61}& 41.65 & N/A \\\hline
 & \multirow{2}{*}{XGB} & \multirow{2}{*}{74.6} & \multirow{2}{*}{Tree} & \multirow{2}{*}{68.6} & KD & 68.8& \textbf{71.2} & 68.80 & 69.40\\\cline{6-10} 
Credit &&&&& PW & 68.9& \textbf{71.4} & 68.8 & N/A \\\cline{2-10}
Card &\multirow{2}{*}{NN} & \multirow{2}{*}{71.8} & \multirow{2}{*}{Tree} & \multirow{2}{*}{68.6} & KD & 64.8 & \textbf{66.40} & 65.20 & 65.80\\\cline{6-10} 
&&&&& PW & 69 & \textbf{69.7} & 68.24 & N/A \\\hline
\multirow{4}{*}{Magic} & \multirow{2}{*}{XGB} & \multirow{2}{*}{85.64} & \multirow{2}{*}{Tree} & \multirow{2}{*}{76.8} & KD & 76.92 & \textbf{77.56} & 72.92 & 73.96\\\cline{6-10} 
&&&&& PW & 76.92& \textbf{77.24} & 76.76 & N/A\\\cline{2-10}
&\multirow{2}{*}{NN} & \multirow{2}{*}{83.4} & \multirow{2}{*}{Tree} & \multirow{2}{*}{76.8} & KD & 72.08 & \textbf{75.52} & \textbf{75.32} & \textbf{75.28}\\\cline{6-10} 
&&&&& PW & 76.6 & \textbf{76.96} & 75.96 & N/A\\\hline
\multirow{4}{*}{Higgs} & \multirow{2}{*}{XGB} & \multirow{2}{*}{70} & \multirow{2}{*}{Tree} & \multirow{2}{*}{59.6} & KD & 57.6 & \textbf{65.8} & 57.80 & 58\\\cline{6-10} 
&&&&& PW & 58.2 & \textbf{62.2} & 58 & N/A\\\cline{2-10}
&\multirow{2}{*}{NN} & \multirow{2}{*}{65.8} & \multirow{2}{*}{Tree} & \multirow{2}{*}{59.6} & KD & 53.4 & \textbf{60.20} & 54.60 & 55.80 \\\cline{6-10} 
&&&&& PW & 58.2 & \textbf{61.8} & \textbf{61.4} & N/A\\\hline
\multirow{4}{*}{Waveform} & \multirow{2}{*}{XGB} & \multirow{2}{*}{85.6} & \multirow{2}{*}{Tree} & \multirow{2}{*}{70.32} & KD & 71.28 &\textbf{72.48} & 70.32 & 70.24 \\\cline{6-10} 
&&&&& PW & 71.08&\textbf{72.32} & 71.28 & N/A\\\cline{2-10}
&\multirow{2}{*}{NN} & \multirow{2}{*}{85.12} & \multirow{2}{*}{Tree} & \multirow{2}{*}{70.32} & KD & 69.12 & \textbf{70.56} & 68.72 & 68.64\\\cline{6-10} 
&&&&& PW & 67.20 & \textbf{68.24} & \textbf{68.4} & N/A\\\hline
\multirow{4}{*}{Diabetes} & \multirow{2}{*}{XGB} & \multirow{2}{*}{77.6} & \multirow{2}{*}{Tree} & \multirow{2}{*}{75} & KD & 76.04 & \textbf{78.12} & 77.04& 76.72\\\cline{6-10} 
&&&&& PW &76.04& \textbf{76.56} & 72.39 & N/A\\\cline{2-10}
&\multirow{2}{*}{NN} & \multirow{2}{*}{81.77} & \multirow{2}{*}{Tree} & \multirow{2}{*}{75} & KD & 71.56 & \textbf{73.96} & 66.67 & 61.46\\\cline{6-10} 
&&&&& PW & 76.04 & \textbf{81.25} & 80.73 & N/A\\\hline
\end{tabular}
\label{tab:realperf}
\end{center}
\end{table*}

\subsection{Real Data Experiments }
\noindent\textbf{Setup:} We experimented on six real datasets (CIFAR100 \cite{cifar} and 5 UCI datasets \cite{uci}) whose characteristics are given in Table \ref{tab:data}. All datasets except CIFAR100 which comes with its own train/test partition were randomly split 10 times into 75\% train and 25\% test. Results were averaged over these random splits. 500 examples from the train set were used for validation and maximum hops $m=3$ for all datasets.

For CIFAR100 the complex model was DenseNet169 ($\approx$ 15M parameters) and the simple model was a LeNet \cite{lenet} ($\approx$ 60K parameters). There were 20 anchor models in this case (i.e. $k=20$) which were smaller DenseNets (DNets) with at least 2M parameters, smaller ResNets, as well as MobileNetV2 ($\approx$ 3.4M parameters). Intermediate models were created from these by following the steps in the \textit{Intermediate Models} section above. The maximum number of perturbations allowed were set to three. For the first hop for DenseNets, the number of repetitions of the Convolution-Relu-BatchNorm sub-block were (randomly) chosen from $\{6, 8, 10, 12, 16, 24\}$ and the growth rate from $\{4, 8, 16, 24, 32\}$. For MobileNetV2, the $\alpha$ was (randomly) chosen between $1$ to $8$
and the number of times a sub-block should be repeated was between $1$ to $5$. These choices were made so that the first hop intermediate model is of lesser complexity than the complex model. For subsequent hops, as mentioned before, these parameters changed depending on the intermediate model chosen in the previous hop. ADAM optimizer was used to train these deep models. The complex model and simple model were trained for 240 epochs, while each of the intermediate models were trained for 70 epochs. The learning rate was set to 0.01 and weight decay was set to 0.1. For the other datasets the complex model was XGboost (XGB)
with 100 trees where each tree had a maximum depth eight as well as a fully connected neural network (NN) of depth 10. The simple model was a single (CART) decision tree of maximum depth 3. There were 228 anchor models for XGB (i.e. $k=228$) which were XGB trees and random forests (RFs) of sizes ranging from 3 to 60 with each tree having a maximum depth of six. Intermediate models were created by (randomly) perturbing the tree levels by utmost two. Thus, the search space was large ruling out brute-force search. For the NN the anchor models were smaller depth NNs from depth two to nine (i.e. $k=8$). Intermediate models were created by randomly adding or removing a layer. $n$ was set to 20 for the XGB and DNet169 experiments. It was set to 8 for the NN experiments as there were only 8 anchor models (i.e. $n=k=8$).

We experimented with two state-of-the-art 1-hop knowledge transfer strategies that work of the original input representations namely, KD and ProfWeight (PW) \cite{profweight}. For DenseNet169 we used the KD code provided here \url{https://github.com/HobbitLong/RepDistiller/}. Since, KD is mainly used to transfer knowledge from neural networks we use its special case for XGB which is model compression \cite{modelcompr}.
Similarly, PW is also primarily applicable to complex models that are neural networks and so we use its special case for XGB where the confidence outputted for the true class serves as the example weight. For baselines we compare with direct training, 1-hop transfer, our method with no perturbations (i.e. $e(.)$ is identity) and an enhanced version of the teaching assistant based knowledge distillation (TAKD) method \cite{takd}, which we refer to as TAKD+. We had to enhance the method as its original implementation requires access to multiple complex models which are neural networks of varying depth with the same architecture and where the simple model also has the same architecture, but is shallower. The method however, is designed for KD as the 1-hop transfer method and is not suited to weighting based transfer methods, which is why we have "N/A" in Table \ref{tab:realperf}. The anchor models were provided to this method as teaching assistants.

\noindent\textbf{Performance:} In Table \ref{tab:realperf} we see the test performance of the different methods. We see that our multihop approach MSTM produces consistently the best results over standard training and 1-hop approaches. The improvement over standard training on the original dataset is on average 3\% and goes up to 6\% in a particular case. The improvement over 1-hop is on average a little more than 2\% and goes above 8\% at best. Best results on CIFAR100 were seen where perturbed versions of DenseNet121 and MobileNetV2 were the intermediate models. In the case of XGB, the best intermediate models varied per dataset and transfer method. In some cases smaller XGBs gave the best results, while in other cases smaller RFs were preferred. Typically, best results were obtained within two to three hops.

\noindent\textbf{Simple model optimized loss vs Evaluation accuracy:} In Table \ref{tab:lossanalysis} we see the losses optimized by the respective simple models. LeNet optimizes cross entropy loss, while (CART) decision tree optimizes Gini index \cite{cart}. In the table to represent the latter as a global loss we compute the weighted sum of the Gini indices of the leaves, where the weights correspond to fraction of examples in the test set that belong to a particular leaf. More formally, if we have a decision tree with $l$ leaves where $\text{GI}_i$ is the Gini index and $n_i$ is the number of examples in the test set for leaf $i$ with $n_t$ being the total number of test examples then, 
$\text{WGI} = \frac{1}{n_t}\sum_{i=1}^{l}n_i\text{GI}_i.$
This measure tells us how well the simple model performs w.r.t. Gini index on the test set.
\begin{table}[h]
\centering
\small
\caption{Below we see the (average) values of the (optimized) losses by the simple models (i.e. cross-entropy for LeNet and (weighted) Gini indices for decision tree) over the test set using KD and using PW as the 1-hop transfer method and XGB as complex model for the tree. Looking at these results in conjunction with Table \ref{tab:realperf} provides more evidence in support of our hypothesis that training and evaluation loss mismatch further motivates multihop.}
\begin{tabular}{|c|c|c|c|c|c|}
  \hline
    \textbf{Dataset} &  \textbf{Loss} & \textbf{Trans. Meth.} & \textbf{SM} & \textbf{1-hop} & \textbf{MSTM}\\
 \hline\hline
 \multirow{2}{*}{CIFAR100} & Cross & KD & {2.47}  & {3.24} & {3.25}\\
 \cline{3-6}
 & entropy& PW & 2.22 & 2.22 & 2.47\\
     \hline
     \multirow{2}{*}{Credit Card} & \multirow{2}{*}{WGI} & KD & 0.38& 0.32&0.06  \\
     \cline{3-6}
     && PW & 0.38 & 0.33 & 0.26\\
     \hline
    \multirow{2}{*}{Magic} & \multirow{2}{*}{WGI} & KD & 0.31 & 0.26 & 0.21 \\
    \cline{3-6}
    && PW & 0.31 & 0.26 & 0.21\\
 \hline
   \multirow{2}{*}{Higgs} & \multirow{2}{*}{WGI} & KD &0.43 & 0.35 & 0.23 \\
   \cline{3-6}
   && PW & 0.43 & 0.40 & 0.38\\
   \hline
 \multirow{2}{*}{Waveform} & \multirow{2}{*}{WGI} & KD & 0.37 & 0.36 & 0.27 \\
 \cline{3-6}
 && PW & 0.37 & 0.35 & 0.31\\
 \hline
\multirow{2}{*}{Diabetes} & \multirow{2}{*}{WGI} & KD & 0.30 & 0.27 & 0.09 \\
\cline{3-6}
&& PW & 0.30 & 0.25 & 0.20\\
 \hline
   \end{tabular}
  \label{tab:lossanalysis}
\end{table}

Looking at Table \ref{tab:lossanalysis} in relation to Table \ref{tab:realperf} we realize that lower simple model loss even on the test set may not correspond to higher accuracy in all cases. For instance, cross entropy loss increases as we go from the model that is directly trained on the dataset to 1-hop to our multihop approach. However, these losses are directly proportional to the test accuracies in Table \ref{tab:realperf}. This is also seen when comparing direct training and 1-hop on the Higgs dataset. Given that these are test set results the reason cannot be model overfitting on the training set that produces this behavior, rather we see the effects of loss mismatch which we also witness in the synthetic experiments. The effect being that \emph{multihop can produce better quality models w.r.t. an evaluation loss even though it may be hard to beat the (training) loss optimized by the simple models}. This is a very interesting and subtle point which seems to be important to witness the power of the multihop approach over its traditional 1-hop counterpart. Moreover, such loss mismatches are prevalent in most machine learning applications as the evaluation loss is typically 0-1 loss, however, one rarely trains a model that directly minimizes it (e.g. NNs optimize cross-entropy, LR optimizes log-likelihood). Qualitatively similar results were seen using PW as the 1-hop method.

\begin{figure}[htbp]
  \centering  
      \includegraphics[width=0.49\textwidth]{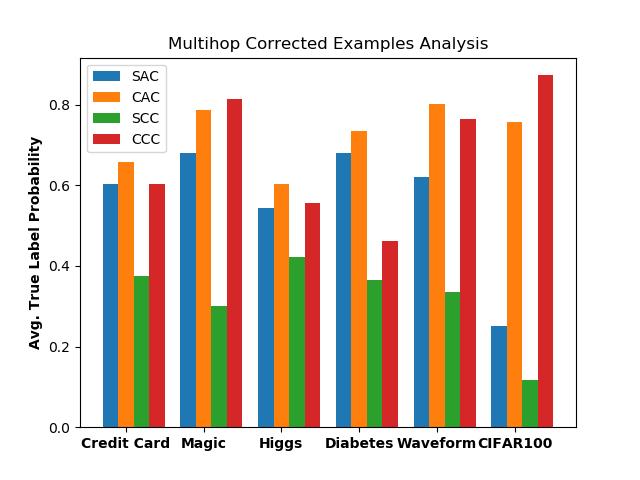}
      \includegraphics[width=0.49\textwidth]{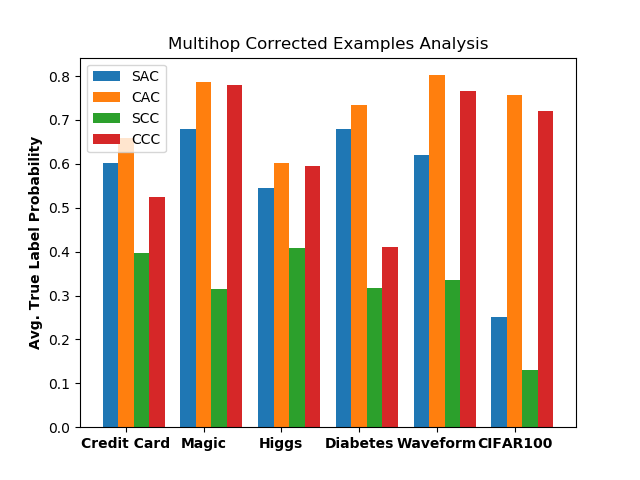}
      
  \caption{Above we see (average) confidences for the true label of the (original) simple model and complex model for two types of test examples. The first is average confidence over the whole test set which are depicted by simple average confidence (SAC) and complex average confidence (CAC). The second is average confidence for examples that were originally misclassified by the simple model but after applying MSTM (using KD (top) and using ProfWeight (bottom)) the new simple model correctly classifies, denoted by simple corrected confidence (SCC) and complex corrected confidence (CCC).}
  \label{multihopptanalysis}
\end{figure}

\noindent\textbf{Type of Examples Corrected by MSTM:} Figure \ref{multihopptanalysis} provides us insight about the type of examples that the simple model trained using MSTM (with KD) correctly classified, but that it
originally misclassified. We see here the relationship between the output confidences of the simple model (and complex model) trained on the dataset for such examples with the average output confidence over all the examples in the test set. In general, it seems that misclassified examples by the simple model that are somewhat near the boundary are targeted for improvement, i.e., they are not confidently misclassified. The simple models confidence for these examples is typically around half its confidence over all examples. The complex model on the other hand is generally confident about these examples, where its confidence on average is similar to its confidence over all examples. Qualitatively similar results were seen with PW as the 1-hop method.
\begin{table}[t]
\small
\centering
\caption{MSTM accuracy restricting anchors or perturbation type (PType) using KD for the different datasets.}
\begin{tabular}{|c|c|c|c|c|c|}
  \hline
    \multirow{2}{*}{\textbf{Dataset}} & \multirow{2}{*}{\textbf{CM}} & \multirow{2}{*}{\textbf{SM}} & \multirow{2}{*}{\textbf{Anchors}} & \multirow{2}{*}{\textbf{PType}} & \textbf{MSTM}  \\
    &&&&& \textbf{Acc}\\
        \hline
        \hline
        \multirow{5}{*}{CIFAR-100} & \multirow{5}{*}{DNet169} & \multirow{5}{*}{LeNet} & DNet & All & 41.36  \\
        \cline{4-6}
         &&& MNetV2 & All & 41.11  \\
        \cline{4-6}
        &&& \multirow{3}{*}{All} & 3 & 42.2\\
        \cline{5-6}
        &&&& 2 & 41.16\\
        \cline{5-6}
         &&&& 1 & 40.55\\
         \hline
         \multirow{4}{*}{Credit card} & \multirow{4}{*}{XGB} & \multirow{4}{*}{Tree} & XGB & All & 69.8  \\
        \cline{4-6}
        &&& RF & All & 70.6\\
        \cline{4-6}
        &&& All & Depth & 69.6\\
        \cline{4-6}
        &&& All & Width & 69.2\\
        \hline
        \multirow{4}{*}{Magic} & \multirow{4}{*}{XGB} & \multirow{4}{*}{Tree} & XGB & All & 77.36  \\
        \cline{4-6}
        &&& RF & All & 77.44\\
        \cline{4-6}
        &&& All & Depth & 77.32\\
        \cline{4-6}
        &&& All & Width & 77.04\\
        \hline
        \multirow{4}{*}{Higgs} & \multirow{4}{*}{XGB} & \multirow{4}{*}{Tree} & XGB & All & 66.8  \\
        \cline{4-6}
        &&& RF & All & 61.6\\
        \cline{4-6}
        &&& All & Depth & 61.8\\
        \cline{4-6}
        &&& All & Width & 60.2\\
        \hline
        \multirow{4}{*}{Waveform} & \multirow{4}{*}{XGB} & \multirow{4}{*}{Tree} & XGB & All & 71.52  \\
        \cline{4-6}
        &&& RF & All & 70.96\\
        \cline{4-6}
        &&& All & Depth & 71.28\\
        \cline{4-6}
        &&& All & Width & 71.2\\
        \hline
        \multirow{4}{*}{Diabetes} & \multirow{4}{*}{XGB} & \multirow{4}{*}{Tree} & XGB & All & 77.6  \\
        \cline{4-6}
        &&& RF & All & 76.6\\
        \cline{4-6}
        &&& All & Depth & 76\\
        \cline{4-6}
        &&& All & Width & 76.6\\
        \hline
  \end{tabular}
 \label{tab:ablation}
\end{table}

\noindent\textbf{Ablations by fixing Anchors or Perturbations:} In Table \ref{tab:ablation} we see more ablation experiments where we either restrict the type of anchors or restrict the perturbations. For example, for CIFAR-100 we show results using just DenseNet (DNet) anchors or MobileNetV2 (MNetV2) anchors, or where we use all the anchors but restrict the number of perturbations to be either one, two or three. Similarly, for the other datasets using XGB as the complex model we restrict to using just smaller ensembles of XGBs (described before) or RFs. We do ablations with XGBs rather than NNs since they were largely better performing. Here the perturbations are restricted to only varying depth or varying only the ensemble size (a.k.a. width). As can be seen from all these experiments that such restrictions lead to inferior performance when compared with Table \ref{tab:realperf}. Nonetheless, looking at the results in Table \ref{tab:ablation} it seems that using all the anchors is more important for CIFAR-100, while interestingly using all the perturbations available is more important for the other datasets. This may be because the architectures of DenseNet and MobileNetv2 are quite different, while we already have enough (diverse) trees using either XGB or RF, where the two sets of anchors can be somewhat redundant. Qualitatively similar results were seen with PW as the transfer scheme.

\section{Discussion}
\label{sec:disc}
In the previous sections we observed and analyzed the behavior of the different approaches. It seems one of the key aspects for effective knowledge transfer to simple models is to have the right kind of function smoothing. For instance, KD increases entropy of the prediction surface, that is, flattens peaks of the class conditional distribution of the complex model. Weighting based approaches such as PW and SRatio remove certain peaks by setting weights of those examples to zero in the extreme case. Multihop seems to perform this smoothing better, where one can in principle use a combination of KD and weighting based methods to reap benefits of both types of smoothening.

In terms of computational complexity the multihop approach is of course more expensive than the 1-hop approach as more alternatives need to be tested. However, given our stochastic scheme the search is over a much smaller space with it still having some performance guarantees as seen in lemma \ref{lem1} over exhaustively searching over the entire state space. Moreover, given the gain in performance the additional complexity may be justified. As such, depending on the available resources one could decide how many models (i.e. $n$)
to test in each hop, as well as how many hops to try. Based on our experience the best models are obtained with two to three hops so the main question is really the value of $n$. Additionally, since at every successive hop our candidate anchor models are only those with lower complexity than the previously chosen anchor model the set of possible candidates can reduce sharply. Also as previously stated, our method seems to be much more efficient than its closest competitor TAKD \cite{takd} which has a time complexity of $O(mk^2)$ as opposed to $O(mn)$ where $n\le k$ which is ours.

In this work we added intermediate models in a sequence, but it may be possible to extend these multihop structures to generic directed acyclic graphs (DAGs). This would add another layer of complexity where one would now have to decide how to aggregate or effectively compute the information to transfer from one hop to the next. Another possible extension is to allow for the option to train intermediate models on arbitrary representations as they do not have to respect the constraints of the final simple model, which may have to be interpretable. For knowledge transfer to such intermediate models it would be interesting to see if ideas from \cite{crd} could be leveraged.


In summary, we have provided a novel meta-approach to transfer knowledge from a complex black-box to a simple model of
predetermined complexity and where there may be also constraints on the representation used to train the simple model (viz. input representation for interpretability). We saw that the approach consistently performed best across different datasets and complex-simple model combinations. An interesting insight that was uncovered was that the multihop approach is especially powerful when there is a mismatch between the loss that the simple model optimizes (viz. log-likelihood) and the evaluation loss (viz. 0-1 loss), which anyway is typically the case in practice.
\bibliography{main_arxiv}

\begin{thebibliography}{10}

\bibitem{weaksubprop}
Y.~Y. Allan~Borodin, Dai Tri Man~Le.
\newblock {Weakly Submodular Functions}.
\newblock {\em arXiv:1401.6697}, 2014.

\bibitem{modelcompr2}
L.~J. Ba and R.~Caurana.
\newblock Do deep nets really need to be deep?
\newblock {\em CoRR}, abs/1312.6184, 2013.

\bibitem{bastani2017interpreting}
O.~Bastani, C.~Kim, and H.~Bastani.
\newblock Interpreting blackbox models via model extraction.
\newblock {\em arXiv preprint arXiv:1705.08504}, 2017.

\bibitem{cart}
L.~Breiman, J.~Friedman, C.~J. Stone, and R.~Olshen.
\newblock {\em Classification and Regression Trees}.
\newblock Chapman and Hall, 1984.

\bibitem{modelcompr}
C.~Bucilu\v{a}, R.~Caruana, and A.~Niculescu-Mizil.
\newblock Model compression.
\newblock In {\em Proceedings of the 12th ACM SIGKDD International Conference
  on Knowledge Discovery and Data Mining}, 2006.

\bibitem{weaksubInit}
A.~Das and D.~Kempe.
\newblock {Submodular meets Spectral: Greedy Algorithms for Subset Selection,
  Sparse Approximation and Dictionary Selection}.
\newblock In {\em Intl. Conference on Machine Learning (ICML)}, 2011.

\bibitem{dehghani2017fidelity}
M.~Dehghani, A.~Mehrjou, S.~Gouws, J.~Kamps, and B.~Sch{\"o}lkopf.
\newblock Fidelity-weighted learning.
\newblock {\em arXiv preprint arXiv:1711.02799}, 2017.

\bibitem{uci}
D.~Dheeru and E.~Karra~Taniskidou.
\newblock {UCI} machine learning repository, 2017.

\bibitem{sratio}
A.~Dhurandhar, K.~Shanmugam, and R.~Luss.
\newblock Enhancing simple models by exploiting what they already know.
\newblock {\em Intl. Conference on Machine Learning (ICML)}, 2020.

\bibitem{profweight}
A.~Dhurandhar, K.~Shanmugam, R.~Luss, and P.~Olsen.
\newblock Improving simple models with confidence profiles.
\newblock {\em Advances of Neural Inf. Processing Systems (NeurIPS)}, 2018.

\bibitem{frosst2017distilling}
N.~Frosst and G.~Hinton.
\newblock Distilling a neural network into a soft decision tree.
\newblock {\em arXiv preprint arXiv:1711.09784}, 2017.

\bibitem{furlanello2018born}
T.~Furlanello, Z.~C. Lipton, M.~Tschannen, L.~Itti, and A.~Anandkumar.
\newblock Born again neural networks.
\newblock {\em arXiv preprint arXiv:1805.04770}, 2018.

\bibitem{distill}
J.~D. Geoffrey~Hinton, Oriol~Vinyals.
\newblock Distilling the knowledge in a neural network.
\newblock In {\em https://arxiv.org/abs/1503.02531}, 2015.

\bibitem{calib}
C.~Guo, G.~Pleiss, Y.~Sun, and K.~Weinberger.
\newblock On calibration of modern neural networks.
\newblock {\em Intl. Conference on Machine Learning (ICML)}, 2017.

\bibitem{resnet}
K.~He, X.~Zhang, S.~Ren, and J.~Sun.
\newblock Deep residual learning for image recognition.
\newblock In {\em Intl. Conference on Computer Vision and Pattern Recognition
  (CVPR)}, 2015.

\bibitem{densenet}
G.~Huang, Z.~Liu, L.~V.~D. Maaten, and K.~Q. Weinberger.
\newblock Densely connected convolutional networks.
\newblock In {\em 2017 IEEE Conference on Computer Vision and Pattern
  Recognition (CVPR)}, pages 2261--2269, 2017.

\bibitem{weaksubstocastic}
R.~Khanna, E.~Elenberg, A.~G. Dimakis, S.~Negahban, and J.~Ghosh.
\newblock {Scalable Greedy Feature Selection via Weak Submodularity}.
\newblock In {\em Intl. Conference on artificial intelligence and statistics
  (AISTATS)}, 2017.

\bibitem{cifar}
A.~Krizhevsky.
\newblock Learning multiple layers of features from tiny images.
\newblock In {\em Tech. Report}. 2009.

\bibitem{lenet}
Y.~Lecun, L.~Bottou, Y.~Bengio, and P.~Haffner.
\newblock Gradient-based learning applied to document recognition.
\newblock In {\em Proceedings of the IEEE}, pages 2278--2324, 1998.

\bibitem{priv16}
D.~Lopez-Paz, L.~Bottou, B.~Sch\"{o}lkopf, and V.~Vapnik.
\newblock Unifying distillation and privileged information.
\newblock In {\em International Conference on Learning Representations (ICLR
  2016)}, 2016.

\bibitem{takd}
S.-I. Mirzadeh, M.~Farajtabar, A.~Li, N.~Levine, A.~Matsukawa, and
  H.~Ghasemzadeh.
\newblock Improved knowledge distillation via teacher assistant.
\newblock In {\em AAAI}, 2020.

\bibitem{molnardecomp}
C.~Molnar, G.~Casalicchio, and B.~Bischl.
\newblock Quantifying interpretability of arbitrary machine learning models
  through functional decomposition.
\newblock {\em CoRR}, 2019.

\bibitem{ren2018learning}
M.~Ren, W.~Zeng, B.~Yang, and R.~Urtasun.
\newblock Learning to reweight examples for robust deep learning.
\newblock {\em arXiv preprint arXiv:1803.09050}, 2018.

\bibitem{fitnet}
A.~Romero, N.~Ballas, S.~E. Kahou, A.~Chassang, C.~Gatta, and Y.~Bengio.
\newblock Fitnets: Hints for thin deep nets.
\newblock {\em arXiv preprint arXiv:1412.6550}, 2015.

\bibitem{mobilenet}
M.~Sandler, A.~Howard, M.~Zhu, A.~Zhmoginov, and L.-C. Chen.
\newblock Mobilenetv2: Inverted residuals and linear bottlenecks.
\newblock In {\em 2018 IEEE Conference on Computer Vision and Pattern
  Recognition (CVPR)}, pages 4510--4520, 2018.

\bibitem{distillnew}
S.~Tan, R.~Caruana, G.~Hooker, and Y.~Lou.
\newblock Auditing black-box models using transparent model distillation with
  side information.
\newblock {\em CoRR}, 2017.

\bibitem{crd}
Y.~Tian, D.~Krishnan, and P.~Isola.
\newblock Contrastive representation distillation.
\newblock In {\em International Conference on Learning Representations}, 2020.

\bibitem{kushwhyInt}
K.~R. Varshney, P.~Khanduri, P.~Sharma, S.~Zhang, and P.~K. Varshney.
\newblock Why interpretability in machine learning? an answer using distributed
  detection and data fusion theory.
\newblock {\em ICML Workshop on Human Interpretability in Machine Learning
  (WHI)}, 2018.

\bibitem{twostageMC}
Z.~Yang, L.~Shou, M.~Gong, W.~Lin, and D.~Jiang.
\newblock Model compression with two-stage multi-teacher knowledge distillation
  for web question answering system.
\newblock In {\em WSDM}, 2020.

\end{thebibliography}
\bibliographystyle{abbrv}
\end{document}